\documentclass[11pt]{article}

\usepackage[T1]{fontenc}
\usepackage[utf8]{inputenc}
\usepackage[margin=1in]{geometry}
\usepackage{booktabs}
\usepackage{multirow}
\usepackage{array}
\usepackage{amsmath}
\usepackage{graphicx}
\usepackage{hyperref}
\usepackage{xcolor}
\usepackage{caption}
\usepackage{enumitem}
\usepackage{authblk}

\hypersetup{colorlinks=true, linkcolor=blue, citecolor=blue, urlcolor=blue}

\newcommand{\code}[1]{\texttt{#1}}

\title{Evaluating Escalation Signals for LLM Routing:\\
Targets, Controls, and Five Ways to Fool Yourself}
\author{Ramin Pishehvar, Andrea Morandi, and Mahesh Viswanathan}
\affil{Cisco Systems, Inc., San Jose, CA, USA \\ \texttt{\{rpishehv, amoranid, mahviswa\}@cisco.com}}
\date{}

\begin{document}
\maketitle

\begin{abstract}
Deciding when to escalate a query from a small language model to a larger one requires a cheap signal that can predict, before the large model is ever called, whether escalating would actually help. Semantic entropy, originally developed to detect hallucinations, is a natural candidate: it measures how much a model's own sampled answers disagree in meaning, and high disagreement often signals an unreliable answer. We test it as an escalation signal across three benchmarks and two model families. On GSM8K, with a small/large model pair about twelve times apart in size, semantic entropy reliably distinguishes the small model's mistakes (AUROC 0.871) and improves routed accuracy over random escalation by up to nine points at matched cost. An earlier, similarly strong-looking result on a synthetic arithmetic benchmark turned out to be misleading: a simple rule based only on how hard a question looks on the page, with no model involved at all, matched semantic entropy's performance almost exactly.

This paper's main contribution is a set of checks that catch this kind of misleading result before it is reported as real. We show that scoring a cheap, question-only difficulty estimate alongside any signal reveals whether that signal is adding real information or just tracking how hard a question looks; that two reasonable-seeming definitions of "the escalation worked" can produce very different results on the same data; that a benchmark can leave almost no room for any signal to beat simply always using the large model, however good that signal is; and that the true cost of gathering a live signal, querying the small model many times per question, can make routing more expensive than just calling the large model directly. For a cheaper alternative that reuses cached past outcomes instead of live sampling, we show how to predict, before running an experiment, whether that shortcut will work on a new dataset, confirmed by correctly forecasting a collapse from a strong-looking result (AUROC 0.908) to a chance-level one (0.518) ahead of time. We offer these as a general checklist for evaluating escalation signals, not as a case for this one.
\end{abstract}

\section{Introduction}

Serving every query with the largest available language model is wasteful: many queries
are easy enough for a small model (SLM) to answer correctly, and only a minority need the
extra capacity of a large model (LLM). A \emph{router} that can cheaply predict, before
querying the large model, whether a given query needs it would let a system pay the large
model's cost only when it is worth paying.

Semantic entropy \cite{farquhar2024detecting} is a natural candidate signal. Originally
proposed for hallucination detection, it estimates a model's uncertainty about the
\emph{meaning} of its own answer: draw $k$ stochastic samples for a query, cluster them by
bidirectional NLI entailment (two samples belong to the same cluster iff each entails the
other), and compute the Shannon entropy of the resulting cluster-probability distribution.
High semantic entropy indicates that the model's samples disagree on meaning, not just
wording -- a stronger hallucination signal than lexical diversity or output confidence
alone. Our implementation clusters a terse \emph{extracted} answer rather than the raw
generation (Section~\ref{sec:bugs}) and takes plain cluster frequency rather than weighting
by the model's own sequence probabilities as in \cite{farquhar2024detecting}; this is
closer to the discrete, frequency-weighted self-consistency of \cite{wang2023selfconsistency}
than to full semantic entropy, and we use the term throughout to mean this specific, cheaper
variant.

This paper asks: can semantic entropy, computed from the \emph{small} model's own samples,
predict whether escalating a query to a \emph{large} model would improve the answer? We
build a full pipeline to test this, run it across three benchmarks and two model families,
and report both positive and negative results, several real implementation bugs
encountered along the way, and a conceptual clarification of what semantic entropy can and
cannot be expected to predict.

\section{Related Work}
\label{sec:related}

\paragraph{Cascades and routing.} Cascading -- serving most queries with a cheap model and
escalating hard ones -- was popularized by FrugalGPT \cite{chen2024frugalgpt}, which learns
confidence thresholds over a portfolio of commercial APIs and reports large cost reductions
at matched quality. AutoMix \cite{aggarwal2024automix} wraps few-shot self-verification in a
POMDP-based router requiring only API access. A parallel line performs \emph{pre-generation}
routing, selecting one model before any inference: RouteLLM \cite{ong2025routellm} and
Hybrid LLM \cite{ding2024hybrid} learn such a router from preference data, the latter
adding an explicit quality-cost objective. The distinction matters for our cost analysis: a
pre-generation router never pays the small model, whereas a cascade pays it on every query
-- and, in our setting, pays it $k$ times (Section~\ref{sec:breakeven}).

\paragraph{Uncertainty signals for deferral.} Token-level confidence is the established
deferral signal for generative cascades, tracing back to the classical error-reject tradeoff
\cite{chow1970rule}; Gupta et al.\ \cite{gupta2024token} formalize the
Chow-style variants (summed and length-normalized log-probability, and quantiles thereof)
that our \code{avg\_neg\_logprob} instantiates. Semantic uncertainty
\cite{kuhn2023semantic} and its refinement into semantic entropy
\cite{farquhar2024detecting} were introduced for hallucination detection rather than
routing.

\paragraph{Agreement-based deferral.} The closest prior work to ours is Soiffer et al.\
\cite{soiffer2025semantic}, who use \emph{semantic agreement} among an ensemble of distinct
small models as a training-free deferral signal, reporting quality matching a 70B target at
40\% of its compute budget. Related, Kolawole et al.\ \cite{kolawole2025agreement} use
ensemble agreement for cascade deferral and Yue et al.\ \cite{yue2024cascades} use
answer consistency for numerical reasoning. Our setting differs in construction -- $k$
stochastic samples from a \emph{single} model, clustered by bidirectional NLI entailment
with entropy over the cluster distribution, following \cite{farquhar2024detecting} -- but
the underlying intuition, that meaning-level disagreement indicates unreliability, is
shared. We regard the effectiveness of this family as established by that work rather than
as our contribution.

\paragraph{What this paper adds.} Several of our findings corroborate results reached
independently, which we take as evidence that they are properties of the problem rather
than of our implementation. Soiffer et al.\ report that agreement-based deferral degrades
on short-answer closed-book QA, where answers are brief and baseline accuracy is low, and
that token-level confidence overtakes it there. We observe a related difficulty on PopQA,
for a different reason: escalation itself is the wrong intervention there, since both models
lack the needed parametric knowledge rather than disagreeing on it (Section~\ref{sec:popqa})
-- we did not reach an entropy-vs-confidence comparison on PopQA, having abandoned signal
evaluation there once the headroom failure was established. On synthetic arithmetic we find
the mirror image of Soiffer et al.'s result instead: token confidence is the signal that
inverts, falling below chance across four independent measurements (Section~\ref{sec:n96}).
The concern motivating
our per-generation cost accounting -- that $k$-sample estimation changes the serving cost
model -- is precisely why at least one recent cascade system declines to use semantic
entropy at all.

Our contribution is therefore not the signal but the evaluation methodology. To our
knowledge, no prior cascade study reports a \emph{difficulty baseline}: the deferral
literature compares against token-level confidence, random deferral, and learned routers,
but not against what the question text alone already reveals. On our synthetic benchmark
that baseline is free and matches semantic entropy exactly, nullifying a result we had
initially accepted (Section~\ref{sec:n96}); on GSM8K entropy beats it by a wide margin
(Section~\ref{sec:gsm8k}), and the two cases are indistinguishable without the control.
Nor, to our knowledge, does prior work report an oracle-headroom precondition, the
difficulty-\emph{calibrated} variant needed to evaluate retrieval-based routers fairly, or
the observation that on lexically regular benchmarks an embedding space can encode
difficulty so completely ($\rho = 0.98$) that nearest-neighbor retrieval is difficulty
lookup -- a diagnostic that correctly forecast the collapse of an apparent retrieval
result on a benchmark not yet run (Section~\ref{sec:kb}). Selective prediction \cite{geifman2017selective} supplies
the abstention framework these controls operate within.

\section{Method}

\subsection{Semantic entropy}

Given a query $q$ and $k$ stochastic samples $y_1, \ldots, y_k$ drawn from a model at
temperature $>0$, we cluster the samples via bidirectional entailment, following the
semantic-uncertainty construction of \cite{kuhn2023semantic} and its refinement into
semantic entropy by \cite{farquhar2024detecting}, using an NLI model
(\code{microsoft/\allowbreak deberta-large-mnli}~\cite{he2021deberta}): $y_i$ and $y_j$ are placed in the same cluster iff
$P(\text{entail} \mid y_i, y_j) > \tau$ and $P(\text{entail} \mid y_j, y_i) > \tau$, for
threshold $\tau = 0.5$. Writing $C_1, \ldots, C_m$ for the resulting clusters and
$p(C_c) = |C_c| / k$ for the empirical cluster probability, semantic entropy is
\begin{equation}
H = -\sum_{c=1}^{m} p(C_c) \log p(C_c),
\end{equation}
and the normalized form used throughout this work is $H_{\text{norm}} = H / \log k \in [0,
1]$, so that entropy is comparable across different values of $k$.

\subsection{Token-level confidence}

Following the Chow-style formalization of \cite{gupta2024token}, we also record,
as a complementary signal, each sample's average per-token
log-probability under the generating model (via \code{compute\_transition\allowbreak\_scores}),
negated so that higher values indicate lower confidence: $\text{neg\_logprob} =
-\frac{1}{T}\sum_{t=1}^{T} \log P(y_t \mid y_{<t})$ for the representative sample chosen
from the majority cluster. This measures the model's confidence in the tokens it actually
produced, independent of whether other samples agreed with it -- a different notion of
uncertainty from semantic entropy, and one that can, in principle, catch cases where a
model is \emph{consistently} wrong (zero semantic entropy) but still assigns low
probability to its own output.

\subsection{Escalation labels and the three evaluation targets}
\label{sec:targets}

For each query we compute \code{correct\_small} and \code{correct\_large} (exact-match or
fuzzy alias match against ground truth, depending on task). Three binary targets can be
constructed from these, which we refer to throughout as Targets 1--3:

\begin{align}
\text{Target 1: } \text{needs\_escalation} &= (\lnot\,\text{correct\_small}) \land \text{correct\_large}, \\
\text{Target 2: } \text{untrustworthy} &= \lnot\,\text{correct\_small}, \\
\text{Target 3: } \text{large\_success} &= \text{correct\_large}.
\end{align}

We initially evaluated signals only against Target 1, since it is the operationally
relevant question (does escalating actually fix the answer?). However, Target 1 depends on
the \emph{large} model's behavior, which a signal computed purely from the small model's
own samples never observes. We are careful to state this limitation precisely: such a
signal carries no \emph{direct} information about the large model, but it may still carry
\emph{correlational} information, mediated by a shared latent -- query difficulty -- that
influences both models' success. Fixed, parameter-free statistics such as entropy exploit
this correlational pathway weakly, whereas a learned probe on internal representations can
exploit it substantially better (Section~\ref{sec:probe}). Target 2 is the target a
per-model uncertainty signal can fairly be judged against, since it only requires
information the small model's own behavior could in principle encode.

Target 3 (defined above) applies to any signal, live or retrieval-based, but the two differ
in what they can be expected to achieve against it: a live small-model signal has only
indirect, difficulty-mediated access to the large model's behavior
(Section~\ref{sec:n96} shows this is nonzero but modest), whereas a knowledge base built by
profiling queries against the large model (Section~\ref{sec:kb}) has direct access to real
\code{correct\_large} outcomes. We report Targets 1 and 2 throughout and add Target 3
wherever a signal has something to say about it, and argue in Section~\ref{sec:reframing}
that this distinction resolves an apparent contradiction in our results.

A note on Target 3's polarity, since it differs by signal family and is easy to misread.
Entropy-family signals (semantic entropy, token confidence) are \emph{uncertainty} scores: a
higher value is designed to indicate trouble, uniformly across Targets 1 and 2. To preserve
that reading for Target 3 as well, Tables~\ref{tab:gsm8k} and~\ref{tab:n96} score these
signals against $\lnot\text{correct\_large}$ rather than against $\text{correct\_large}$ as
defined above -- without this inversion, a higher-is-better entropy score would correspond to
an AUROC read as $1-\text{AUROC}$ under the literal definition, inviting exactly the kind of
misreading this note exists to prevent. Retrieval-based signals (Section~\ref{sec:kb}) are
different in kind: they vote directly on the outcome rather than report uncertainty, so a
higher score naturally means \emph{predicted success}, and Table~\ref{tab:kb-targets} scores
them against $\text{correct\_large}$ directly, matching the definition above without
inversion. The two conventions are each the correct one for their own signal family; neither
table is in error, but a reader comparing Target 3 AUROC values across both kinds of signal
should note that they are scored against opposite classes.

\subsection{Cost model}
\label{sec:cost}

Every query incurs the cost of the $k$-sample small-model pass, normalized to $1$ unit
(this is unavoidable, since the entropy signal requires it). Escalated queries
additionally incur \code{cost\_ratio} units for one large-model call. For escalation rate
$r$ (fraction of queries escalated) and cost ratio $\rho$,
\begin{equation}
\text{cost}(r) = 1 + \rho \cdot r.
\end{equation}
We compare routed accuracy at a given cost against three baselines: always-small
(cost $1$), always-large (cost $1+\rho$), and \emph{random-at-matched-cost} (escalate a
random subset of the same size $r \cdot n$, averaged over 20 trials) -- the last of these
isolates whether the signal's escalation choices are actually better than an
equal-cost random policy, rather than merely benefiting from escalating more often.
We also compare against an oracle, in the spirit of the abstention framework of
\cite{geifman2017selective}: a hypothetical router that always makes the correct escalate-or-not decision. Its accuracy is capped by how many queries the small model already answers correctly that the large model would get wrong, since those are the only queries where escalating can actively hurt. This gives a cheap pre-flight check: an oracle's margin over always-large bounds what any router, however good its signal, could possibly gain, so a small margin means a benchmark is not worth a full evaluation.

\subsection{The sampling overhead this cost model hides}
\label{sec:breakeven}

Normalizing the $k$-sample pass to one unit obscures a constraint that turns out to bind
hard in practice: entropy routing must draw $k$ generations from the small model
\emph{before} it can decide anything, and it pays that cost on every query, including the
ones it does not escalate. Charging per generation instead, with a large-model generation
costing $\gamma$ times a small-model one,
\begin{equation}
\text{cost}_{\text{router}} = k + r\gamma, \qquad
\text{cost}_{\text{always-large}} = \gamma,
\end{equation}
so routing is cheaper than simply always calling the large model only when
\begin{equation}
\gamma > \frac{k}{1 - r}.
\label{eq:breakeven}
\end{equation}
At $k=10$ this requires $\gamma > 11.1$ even at a $10\%$ escalation rate, and
$\gamma > 14.3$ at $30\%$. The consequence is that the method's viability depends on the
model pair, not only on signal quality: a 270M/12B pair ($\sim$44$\times$ parameters)
plausibly clears the bar, whereas the 1B/12B pair used in Section~\ref{sec:popqa}
($\sim$12$\times$) does not, and for that pair entropy routing is more expensive than
always-large no matter how well the signal discriminates. We note this because it is easy to
miss under a normalized cost model, and because it bounds the practical claim this line of
work can make: semantic-entropy routing is only economically sensible at large capability
(and hence large cost) ratios, or at substantially smaller $k$ than we use here. In
Section~\ref{sec:kb}, we propose an offline method based on retrieval to make the complexity
argument above more appealing for real-world applications.

\subsection{The difficulty baseline: a necessary control}
\label{sec:diffbaseline}

An escalation signal is only interesting if it beats what is already known about the query
for free. We therefore score, alongside every signal, a baseline that uses the benchmark's
own ground-truth difficulty as the routing score.

This baseline is a \emph{control}, not a candidate router. Its role is the same as the
oracle row in a Pareto table: an oracle is not deployable either, but it bounds what any
router could achieve, and a signal approaching it means something. The question the
difficulty baseline answers is diagnostic -- \emph{does this signal carry information
beyond how hard the question is?} -- and that question needs an answer whether or not
difficulty happens to be available at inference. If the answer is no, one should build a
cheap difficulty estimator rather than pay for $k$ samples and NLI inference.

Whether the baseline is additionally a \emph{deployable} alternative depends on where its
difficulty annotation comes from, and this varies sharply across our benchmarks
(Table~\ref{tab:diffsource}). Two regimes must be distinguished:

\begin{table}[h]
\centering
\caption{Where each benchmark's difficulty annotation comes from, and what the baseline
therefore means. Only on the synthetic ladder is difficulty recoverable from the question
at inference time; elsewhere the baseline is a skyline using privileged information.}
\label{tab:diffsource}
\resizebox{\textwidth}{!}{%
\begin{tabular}{llll}
\toprule
Benchmark & Difficulty source & Available at inference? & Baseline is a \\
\midrule
Synthetic & digit count $\times$ operation count, from the question & \textbf{yes} (regex) & free competitor \\
GSM8K & calculation lines in the \emph{reference solution} & no (needs the answer) & skyline \\
PopQA & entity popularity, dataset field & no (external lookup) & skyline \\
\bottomrule
\end{tabular}%
}
\end{table}

\begin{itemize}
\item When difficulty is a \emph{surface feature} of the question -- as on our synthetic
arithmetic ladder, where it is operand digit count times operation count and a regex
recovers it exactly -- the baseline is a genuine zero-cost competitor. A signal that does
not beat it provides no value, because the same routing decisions are available without any
model call at all.
\item When difficulty is \emph{latent} -- derived from the reference answer (GSM8K's
reasoning-step count) or requiring an external lookup (PopQA's entity popularity) -- the
baseline is instead a \emph{skyline}: it uses information unavailable at inference, since
knowing how many reasoning steps a problem takes presupposes the solution one is trying to
obtain. Beating it is then a strong result rather than a null one.
\end{itemize}

This control is cheap and, in our experience, decisive: it converted our strongest
synthetic result from a positive finding into a null one (Section~\ref{sec:n96}),
and it is what distinguishes that null from the genuine GSM8K result
(Section~\ref{sec:gsm8k}). We recommend it be reported by default. Its absence is, we
suspect, why difficulty-correlated signals are easy to over-credit: any signal that tracks
difficulty will separate a difficulty-driven target, whether or not it contains
query-specific information.

We report two variants of this baseline, and recommend taking the larger of the two as the
bar a signal must clear. The distinction turns on an assumption that scoring difficulty
directly makes silently.

\paragraph{The raw baseline assumes the target is monotone in difficulty.} AUROC measures
ranking quality, so using difficulty itself as the routing score asserts that higher
difficulty implies a higher probability of the positive label. Where that assumption fails,
the baseline scores poorly \emph{even when difficulty is highly informative}, and a signal
compared only against it can appear to add information it does not have.

Target 1 is the clearest case, because its positives are concentrated in the middle of the
difficulty range rather than at the top. Writing the three regimes out:

\begin{center}
\begin{tabular}{llll}
\toprule
Difficulty & \code{correct\_small} & \code{correct\_large} & Target 1 \\
\midrule
low    & yes & yes & negative \\
medium & no  & yes & \textbf{positive} \\
high   & no  & no  & negative \\
\bottomrule
\end{tabular}
\end{center}

Escalation helps only where the small model has failed \emph{and} the large model succeeds.
At low difficulty the small model is already correct; at high difficulty both fail. So
$P(\text{Target 1} \mid \text{difficulty})$ is single-peaked, and ranking by raw difficulty
places the hardest items -- which are negatives -- at the top. The ranking is not merely
uninformative but wrong at precisely the end that a router acts on. Empirically this is what
we observe: raw difficulty scores $0.522$ against Target 1 on GSM8K
(Table~\ref{tab:gsm8k}), indistinguishable from chance, despite difficulty plainly bearing
on whether escalation helps.

\paragraph{The calibrated baseline estimates the shape instead of assuming it.} Rather than
scoring difficulty, we score the \emph{empirical rate of the target among items of similar
difficulty}, computed leave-one-out: for each item, we take its nearest neighbors in
difficulty among the remaining items and use the fraction of them that are positive. This
estimates $P(\text{label} \mid \text{difficulty})$ nonparametrically, so a single-peaked
relationship is represented correctly -- medium-difficulty items receive high scores, and
both tails receive low ones. The leave-one-out construction is necessary rather than
fastidious: including an item's own label in its own score would make the score partly
\emph{be} the label.

Concretely, the neighbor count is fixed at $8$, not chosen by cross-validation: for each
item we first collect every other item at the \emph{exact same} difficulty value, and fall
back to the $8$ nearest by absolute difficulty distance only when fewer than $8$ exact
matches exist. This favors exact-match neighbors where the benchmark's difficulty levels are
discrete enough to supply them (as on the synthetic ladder), and degrades gracefully to a
fixed-radius nearest-neighbor estimate otherwise (as on GSM8K, where difficulty values are
less repetitive).

The size of this effect is not merely theoretical, but we are careful here to cite only
what we can verify. Target 1 is single-peaked by the argument above on every benchmark we
test, and raw difficulty scores accordingly poorly against it throughout: $0.522$ on GSM8K
(Table~\ref{tab:gsm8k}), $0.517$--$0.560$ across GSM8K's own $k$-ablation samples
(Table~\ref{tab:kfull}), and $0.559$--$0.563$ on the synthetic benchmark
(Table~\ref{tab:n96}) -- all indistinguishable from chance. We do
not, however, have a verified instance in our own data of the calibrated variant turning
this into a large gain on Target 1 specifically -- GSM8K's calibrated score ($0.499$,
Table~\ref{tab:gsm8k}) is if anything slightly worse than raw there, consistent with our
later point that calibration helps only when the benchmark's difficulty annotation itself
correlates with outcomes, which GSM8K's reasoning-step count does weakly at best. The
clearest demonstrated instance of calibration recovering non-monotone structure in this
paper is therefore the raw-difficulty failure itself, not a specific calibrated figure we
can point to: a baseline that scores nothing against Target 1's non-monotone shape would
need calibration to express it at all, even where we cannot yet show by how much.

\paragraph{Neither variant dominates.} Calibration is not free: it estimates a rate from a
handful of neighbors, and that estimate is noisy. Where the relationship genuinely is
monotone, raw difficulty already expresses it exactly and calibration can only perturb the
ordering. GSM8K Target 2 illustrates this -- $\lnot$\code{correct\_small} rises
monotonically with difficulty, and raw difficulty ($0.649$) accordingly beats the calibrated
variant ($0.558$), which is estimating a shape that was not in doubt. The two variants are
thus complementary: calibration recovers structure the raw form cannot express, and the raw
form avoids estimation noise the calibrated form incurs. Reporting only one invites a signal
to clear whichever bar happens to be lower on that dataset, which is why we take the maximum.

\paragraph{An asymmetry worth stating.} The calibrated baseline consumes other items'
labels, whereas semantic entropy observes none. It is therefore a \emph{fair} comparator for
retrieval-based routers -- a nearest-neighbor vote over cached outcomes is itself a
calibrated conditional estimate, so scoring such a router against raw difficulty alone
systematically overstates it (Section~\ref{sec:kb}) -- and a deliberately \emph{harsh} one
for live signals. Semantic entropy exceeding it by $+0.313$ on GSM8K Target 2 is
correspondingly stronger evidence than the margin alone conveys, since it is outperforming a
predictor granted information the signal itself never receives.

\subsection{Harmful and wasted escalation}
\label{sec:harm}

Beyond aggregate accuracy, we track two per-escalation diagnostics. Both are
\emph{post-hoc evaluation} quantities: they require ground-truth grading of both models and
are therefore not computable at deployment time, but they characterize the failure modes a
routing policy incurs in a way aggregate accuracy obscures:
\begin{align}
\text{harmful} &= P(\text{correct\_small} \land \lnot\,\text{correct\_large} \mid \text{escalated}), \\
\text{wasted} &= P(\text{correct\_small} \mid \text{escalated}).
\end{align}
\code{harmful} captures the worst outcome -- escalation actively breaking a correct
answer; \code{wasted} captures the broader cost of escalating when the small model was
already adequate, regardless of what the large model does.

\section{Implementation}

The pipeline consists of an inference module (\code{InferenceEngine}, wrapping any
HuggingFace causal LM, generating $k$ samples with optional per-token log-probabilities),
a semantic-entropy module (\code{NLIEntailmentChecker} and
\code{SemanticEntropyEstimator}, implementing the clustering and entropy computation
above), an answer-extraction module (\code{AnswerExtractor}, reducing a raw generation to
a terse final answer via numeric extraction or last-sentence extraction before
clustering), a router (\code{Router}, implementing the threshold escalation policy end to
end), and an evaluation harness (\code{evaluate.py}) tying all of the above together
across five benchmark loaders (synthetic arithmetic, GSM8K \cite{cobbe2021gsm8k},
PopQA \cite{mallen2023trust}, TriviaQA \cite{joshi2017triviaqa}, TruthfulQA
\cite{lin2022truthfulqa})
with a pre-flight sanity check and full AUROC/AUPRC/Pareto reporting. This paper reports
results on three of the five (synthetic arithmetic, GSM8K, PopQA); TriviaQA and TruthfulQA
loaders exist in the harness but were not run for this study.

\subsection{Experimental settings}
\label{sec:settings}

Unless stated otherwise, all runs use $k=10$ samples per query, an NLI entailment threshold
of $\tau = 0.5$, a maximum generation length of 256 new tokens (raised from an initial 128
to accommodate chain-of-thought reasoning; see Section~\ref{sec:bugs}), and nucleus
sampling \cite{holtzman2020nucleus} at temperature $0.7$ with $\text{top-}p = 0.95$ and $\text{top-}k = 50$.
Stochastic sampling is required rather than optional: greedy decoding would return $k$
identical samples and yield zero entropy regardless of the model's true uncertainty.
Generation models are loaded in \code{bfloat16}; the NLI model
(\code{microsoft/\allowbreak deberta-large-mnli} \cite{he2021deberta}, $\sim$400M parameters) is kept in \code{float32}
(Section~\ref{sec:bugs}). We later ablate the sample budget $k$
(Section~\ref{sec:kablation}); we did not perform a sensitivity analysis over $\tau$, which
remains a natural candidate for ablation. 

Costs are reported as escalation rate rather than in absolute units. This is
without loss of generality: under the model of Section~\ref{sec:cost}, a budget equal to
fraction $f$ of always-large's premium permits escalation rate $r \le f$ for \emph{any}
$\rho$, since $1 + \rho r \le 1 + f\rho \iff r \le f$. The cost ratio is thus a units
conversion rather than a free parameter, and no sensitivity analysis over it is meaningful.

\paragraph{Reproducibility.} All runs were executed on a single NVIDIA L4 GPU. The
$k$-sample ablation of Section~\ref{sec:kablation} used a fixed seed ($0$) for item
selection across $k\in\{3,5,10\}$, so item selection did not vary across that ablation;
the generations themselves are stochastic (temperature $0.7$) and are not separately
seeded, which is the source of the run-to-run variance discussed in
Section~\ref{sec:n96}. From the timestamps recorded in our run logs, consecutive
$250$-item GSM8K runs at this $k$ range completed roughly $75$--$80$ minutes apart; we
report this as an approximate, retrospective interval between logged runs rather than a
controlled timing measurement, since other processes may have run on the same hardware
between them. Seed values for the GSM8K replication sample and the $96$-item synthetic
re-analysis (Sections~\ref{sec:gsm8k} and~\ref{sec:n96}) were not separately recorded
beyond noting that they used a different or disjoint seed from the primary run; we report
this limitation rather than reconstruct specific values we do not have.

\section{Debugging Log: Real Failure Modes Encountered}
\label{sec:bugs}

Several non-obvious implementation issues surfaced during development, each of which
would silently distort results if left unfixed. We record them here since they recurred
across the project and are not specific to any one experiment.

\begin{description}[leftmargin=0pt, style=nextline]
\item[Stylistic-elaboration confound.] Early raw-generation clustering conflated
  genuine disagreement with stylistic embellishment: for the query ``What is the capital
  of Australia?'', several samples correctly named Canberra but appended fabricated
  tangential details (wrong region, invented history, fictitious statistics). Bidirectional
  entailment correctly failed on these pairs -- the longer answer's extra claims are not
  entailed by the short answer -- inflating entropy to near-maximum (\(H_{\text{norm}} =
  1.0\)) despite near-unanimous agreement on the actual answer. This motivated
  \code{AnswerExtractor}, which clusters a terse extracted answer instead of the raw
  generation.

\item[NLI question-context suppression.] Wrapping each answer in a \code{"Question:
  \ldots\ Answer: \ldots"} template before the NLI check -- intended to disambiguate
  short answers -- turned out to be out-of-distribution for a general-purpose MNLI
  checkpoint, systematically suppressing entailment probabilities and inflating entropy
  even for textually near-identical answers. Disabling this wrapper by default resolved
  the issue; it is now opt-in.

\item[dtype mismatch on the NLI model.] Loading the NLI model in \code{bfloat16} (to save
  memory) produced a \code{RuntimeError} inside DeBERTa's disentangled-attention matmul,
  since some internal buffers do not consistently cast across backends. The fix is to keep
  the (small, $\sim$400M-parameter) NLI model in \code{float32} while using
  \code{bfloat16} only for the (much larger) generation models, where the memory savings
  matter.

\item[OOM during model loading.] Instantiating models with the library default
  (\code{float32}, without \code{low\_cpu\_mem\_usage}) caused a transient peak roughly
  double the model's steady-state memory footprint during the loading step itself,
  triggering an OOM kill even on hardware that could otherwise hold the model comfortably.
  Setting \code{low\_cpu\_mem\_usage=True} and defaulting to \code{bfloat16} resolved
  this.

\item[Prompt-induced capability-gap suppression.] An early arithmetic prompt
  (``Answer with just the final number'') forced single-shot mental arithmetic with no
  scratchpad, for both models. At higher difficulty (multi-digit, multi-step
  expressions), this erased most of a larger model's real advantage, which stems
  substantially from chain-of-thought reasoning (carrying digits, tracking intermediate
  sums). The small and large models then performed similarly poorly, which superficially
  resembled ``the large model is too weak'' but was actually an artifact of a prompt that
  suppressed the mechanism by which model scale helps on this task. Allowing reasoning
  (``Show your work step by step, then state the final answer'') and extracting only the
  last number for grading resolved this.

\item[Lone-surrogate Unicode crash.] Generation occasionally produced invalid UTF-8 (a
  lone surrogate character), typically when \code{max\_new\_tokens} truncated output
  mid-multi-byte-character. This crashed CSV serialization at the very end of a full run.
  Sanitizing decoded text immediately after generation (\code{encode(...,
  errors="replace").decode(...)}) fixes this at the source.

\item[Small-sample AUROC instability.] A promising result at $n=15$ (semantic-entropy
  AUROC against \code{needs\_escalation} of $0.602$--$0.630$ for the combined signal) did
  not replicate at $n=100$ on the same benchmark family (AUROC collapsed to
  $0.50$--$0.53$, chance level), despite the larger run having a more favorable
  escalation-worthy population by every other measure. We return to this in
  Section~\ref{sec:n96}.
\end{description}

\section{Results}

We open with the two benchmarks side by side, because the contrast between them is the
result. Semantic entropy posts a strong-looking score on both; on one it survives the
controls of Section~\ref{sec:diffbaseline} and on the other it does not, and no number
visible without those controls distinguishes the two cases. Sections~\ref{sec:popqa}
and~\ref{sec:n96} then give the failing benchmarks in full, which is where most of the
methodological content lies.

\begin{figure}[h]
\centering
\includegraphics[width=0.85\textwidth]{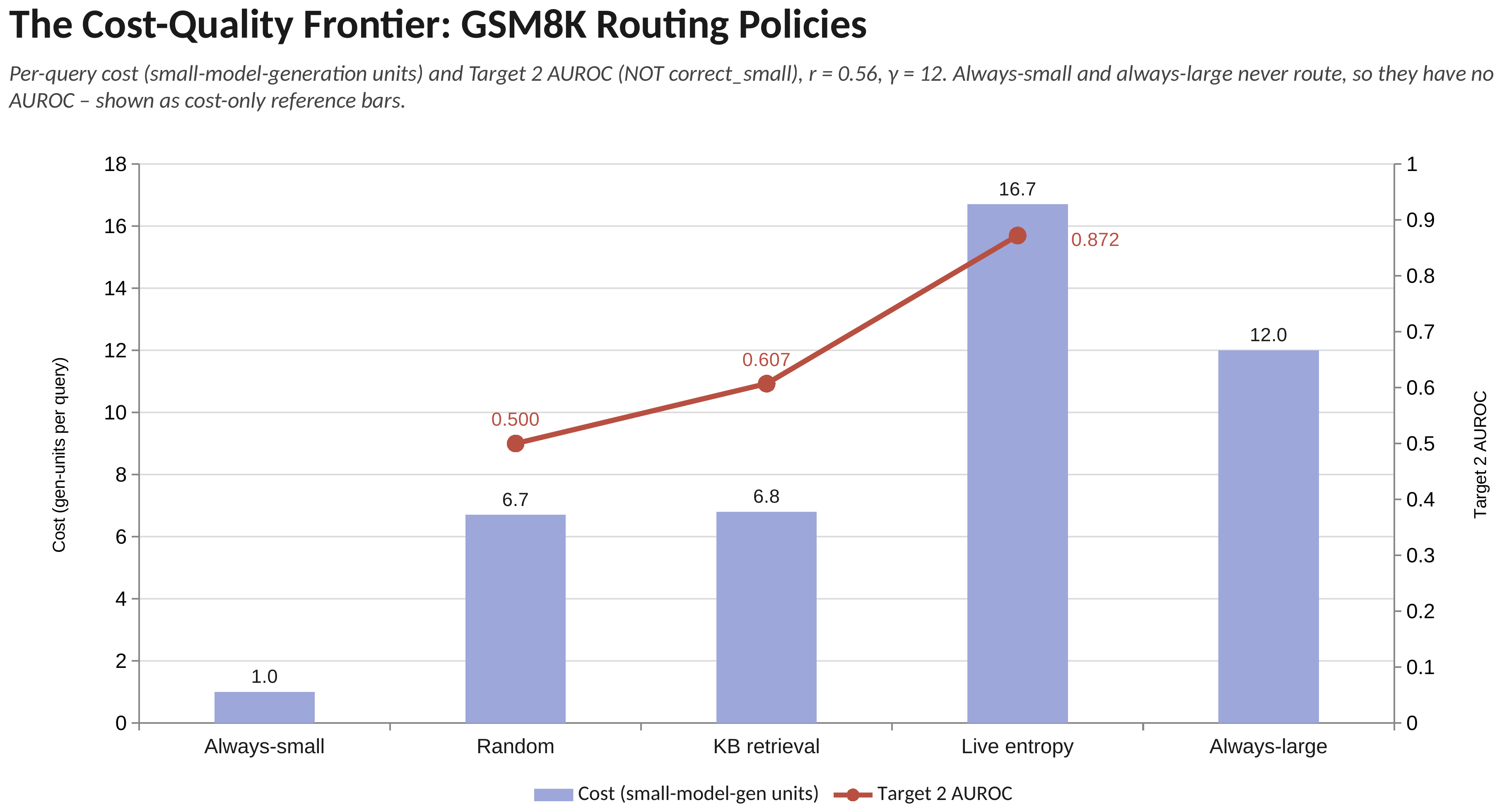}
\caption{The cost-quality frontier on GSM8K ($r = 0.56$, $\gamma = 12$), drawn from
Table~\ref{tab:tradeoff} together with the cost model of Section~\ref{sec:cost} (which
supplies always-small's cost of $1$ unit; always-small has no row in Table~\ref{tab:tradeoff}
itself). Always-small and always-large never route, so they carry no AUROC and are shown as
cost-only reference bars.}
\label{fig:frontier}
\end{figure}

\subsection{Two benchmarks, one signal, opposite verdicts}
\label{sec:gsm8k}

\paragraph{The synthetic benchmark first, because it is the cautionary case.} Our initial
experiments used a synthetic chained-arithmetic ladder with difficulty controlled by operand
digit count and operation count. On it, semantic entropy separates the fair target
($\lnot$\code{correct\_small}) at AUROC $0.830$ and $0.867$ across two profiling runs --
figures we accepted as a positive result. Applying the difficulty baseline of
Section~\ref{sec:diffbaseline} withdraws it. Scoring the benchmark's own difficulty
directly, which on this ladder a regular expression recovers from the question text at no
cost, gives $0.839$ and $0.830$: a tie, with the sign of the difference flipping between
runs and both gaps inside the $0.04$ noise floor. Ten stochastic generations and NLI
clustering purchase no routing decisions a regex does not already supply. A second
precondition fails independently -- oracle headroom over always-large is only $+0.021$, so
even a perfect router had almost nothing to win. Section~\ref{sec:n96} gives the full
analysis; what matters here is that neither failure is visible in the $0.830$ itself.

\paragraph{GSM8K, where the same signal survives the same controls.} We evaluated a Gemma-3
1B (small) / 12B (large) pair \cite{gemma3team2025} on $n=250$ GSM8K \cite{cobbe2021gsm8k} items with $k=10$ samples per query. This
configuration passes the headroom precondition that synthetic arithmetic failed:
always-small accuracy $0.372$, always-large $0.484$, and oracle $0.572$, leaving $+0.088$ of
accuracy headroom -- four times what the synthetic ladder offered. The both-wrong
population is $0.428$.

\begin{table}[h]
\centering
\caption{GSM8K, Gemma-3 1B/12B, $n=250$, $k=10$. AUROC with 95\% bootstrap intervals
\cite{efron1993bootstrap}.
The difficulty baseline uses GSM8K's own difficulty annotation as the routing score.}
\label{tab:gsm8k}
\begin{tabular}{llll}
\toprule
Signal & T1: needs\_escalation & T2: NOT correct\_small & T3: NOT correct\_large \\
\midrule
Semantic entropy      & 0.734 [0.668, 0.799] & \textbf{0.871} [0.820, 0.915] & 0.639 [0.570, 0.709] \\
Token neg-logprob     & 0.648 [0.562, 0.731] & 0.647 [0.577, 0.711] & 0.490 [0.419, 0.561] \\
Combined              & \textbf{0.749} [0.685, 0.813] & 0.852 [0.797, 0.900] & 0.608 [0.536, 0.678] \\
\midrule
Difficulty (raw)      & 0.522 [0.444, 0.603] & 0.649 [0.584, 0.712] & 0.633 [0.564, 0.702] \\
Difficulty-calibrated & 0.499 [0.416, 0.583] & 0.558 [0.486, 0.629] & 0.535 [0.460, 0.610] \\
\bottomrule
\end{tabular}
\end{table}

Three things stand out. First, semantic entropy separates Target 2 at $0.871$, an interval
well clear of chance, and the operationally relevant Target 1 at $0.734$ -- substantially
better than the $0.613$ and $0.666$ the same signal managed on synthetic arithmetic.

Second, and decisively, the comparison that withdrew the synthetic result confirms this one.
Entropy beats raw difficulty by $+0.212$ on Target 1 and $+0.222$ on Target 2, and the
harsher difficulty-calibrated baseline by $+0.235$ and $+0.313$, with disjoint confidence
intervals on Target 2 ($[0.820, 0.915]$ vs.\ $[0.486, 0.629]$). Where the synthetic gap was
$-0.009$ and $+0.037$ -- inside the noise floor, with the sign unstable -- these margins
exceed it by a factor of five and hold in the same direction across both baselines. The two
benchmarks are separated not by the raw AUROC, which is comparable ($0.830$ against
$0.871$), but entirely by the margin over what difficulty already explains. Entropy carries
query-specific information here and did not there.

It is worth being precise about what the baseline had access to on this benchmark, because
it strengthens the claim. GSM8K ships no difficulty label, so we approximate difficulty by
the number of calculation lines in the \emph{reference solution}
(Section~\ref{sec:diffbaseline}). That quantity is derived from the answer: knowing how many
reasoning steps a problem requires presupposes having solved it. The difficulty baseline on
GSM8K is therefore a skyline rather than a deployable competitor, and semantic entropy --
which observes only the small model's own samples, and no labels at all -- outperforms a
predictor that was permitted to inspect the ground-truth solution. The calibrated variant is
stronger still in this respect, since it additionally consumes other items' outcome labels.
A signal beating both is doing something neither difficulty nor difficulty-conditioned
outcome frequency can do: distinguishing between two problems that require the same number
of steps.

Note that on GSM8K the calibrated baseline is \emph{weaker} than raw difficulty ($0.558$
vs.\ $0.649$ on Target 2), the reverse of what we observe on synthetic arithmetic. The two
baselines are not ordered in general: calibration estimates $P(\text{label} \mid
\text{difficulty})$ from a handful of same-difficulty neighbors, which recovers
non-monotonic structure that raw difficulty cannot express, but injects estimation noise
where the relationship is monotone and neighbors are few. GSM8K's difficulty annotation
correlates only weakly with outcomes ($\rho = 0.319$ for entropy against difficulty, versus
$0.657$ on synthetic), so calibration costs more than it recovers. We therefore recommend
reporting both and treating the \emph{maximum} as the bar a signal must clear.

Third, Target 3 is the exception: entropy ($0.639$) and raw difficulty ($0.633$) are
indistinguishable, though entropy does beat the calibrated variant ($0.535$). This is
consistent with the structural argument of Section~\ref{sec:targets} -- whatever a
small-model-only signal knows about the \emph{large} model's success appears to be mediated
by shared query difficulty, with little additional query-specific contribution.

\paragraph{Routing performance.} The signal converts into real routing gains
(Table~\ref{tab:gsm8k-pareto}). Accuracy exceeds the random-at-matched-cost baseline at
every operating point above a $4\%$ escalation rate, peaking at $+0.093$, and exceeds
always-large accuracy for escalation rates between roughly $0.37$ and $0.89$ -- that is,
the router is not merely approximating the large model more cheaply, it is outperforming
it, by declining to escalate items the large model would have gotten wrong.

\begin{table}[h]
\centering
\caption{GSM8K Pareto curve, routing on semantic entropy. ``margin'' is over
random-at-matched-cost; ``vs large'' is over always-large accuracy ($0.484$).}
\label{tab:gsm8k-pareto}
\begin{tabular}{rrrrrrr}
\toprule
Esc.\ rate & Accuracy & Random & Margin & vs large & Harmful & Wasted \\
\midrule
0.00 & 0.372 & 0.372 & $-$0.000 & $-$0.112 & 0.000 & 0.000 \\
0.17 & 0.424 & 0.391 & $+$0.033 & $-$0.060 & 0.023 & 0.093 \\
0.29 & 0.468 & 0.405 & $+$0.064 & $-$0.016 & 0.014 & 0.083 \\
0.37 & 0.500 & 0.413 & $+$0.087 & $+$0.016 & 0.022 & 0.087 \\
0.56 & 0.528 & 0.435 & $\mathbf{+0.093}$ & $+$0.044 & 0.043 & 0.135 \\
0.77 & 0.516 & 0.460 & $+$0.056 & $+$0.032 & 0.073 & 0.207 \\
0.89 & 0.496 & 0.473 & $+$0.023 & $+$0.012 & 0.085 & 0.300 \\
\midrule
oracle & 0.572 & -- & -- & $+$0.088 & 0.000 & 0.000 \\
\bottomrule
\end{tabular}
\end{table}

\paragraph{Replication on an independent sample.} The knowledge-base experiment of
Section~\ref{sec:kb} incidentally profiles a second, disjoint GSM8K sample with the same
model pair ($n=150$, drawn with a different seed), which we can score with the same
signals. It replicates the result above (Table~\ref{tab:gsm8k-replication}). On Target 2 the
two runs agree to within $0.008$ AUROC -- a fifth of the $\approx 0.04$ noise floor of
Section~\ref{sec:n96}, and notably tighter than that floor despite being a harder test,
since the floor is measured across re-samplings of a fixed item set whereas these two runs
draw different items. Entropy's margin over the stronger difficulty baseline is $+0.222$
and $+0.242$ respectively. Target 1 differs more ($0.734$ vs.\ $0.648$), which at these sample sizes we do not regard
as resolvable: the two intervals overlap substantially. The intervals also show the two
targets to be on different footings. On Target 2, entropy's interval is disjoint from both
difficulty baselines' in both samples, so the margin is not a sampling artifact. On Target 1
that separation holds only in the larger sample; at $n=150$ entropy's interval
($[0.549, 0.743]$) overlaps raw difficulty's ($[0.414, 0.607]$), so the Target 1 margin
should be read as directional rather than established.

\begin{table}[h]
\centering
\caption{The GSM8K result on two disjoint samples with the same Gemma-3 1B/12B pair. AUROC
with $95\%$ percentile bootstrap intervals over $2000$ resamples. On Target 2, entropy's
interval is disjoint from both difficulty baselines' in both samples; on Target 1 the
entropy intervals from the two samples overlap substantially, which is why we do not treat
their difference as resolvable. The oracle headroom row is a point estimate and is not
bootstrapped.}
\label{tab:gsm8k-replication}
\begin{tabular}{lcc}
\toprule
& $n=250$ & $n=150$ \\
\midrule
T2: semantic entropy        & \textbf{0.871} [0.820, 0.915] & \textbf{0.863} [0.790, 0.926] \\
T2: difficulty (raw)        & 0.649 [0.584, 0.712] & 0.621 [0.533, 0.708] \\
T2: difficulty (calibrated) & 0.558 [0.486, 0.629] & 0.606 [0.512, 0.705] \\
\midrule
T1: semantic entropy        & \textbf{0.734} [0.668, 0.799] & \textbf{0.648} [0.549, 0.743] \\
T1: difficulty (raw)        & 0.522 [0.444, 0.603] & 0.508 [0.414, 0.607] \\
T1: difficulty (calibrated) & 0.499 [0.416, 0.583] & 0.497 [0.398, 0.602] \\
\midrule
Oracle headroom over always-large & $+0.088$ & $+0.093$ \\
\bottomrule
\end{tabular}
\end{table}

\paragraph{Token confidence is not universally anti-informative.} On GSM8K, neg-logprob
reaches $0.648$/$0.647$ on Targets 1 and 2 -- weak but clearly above chance -- and fusing
it with entropy \emph{improves} Target 1 ($0.749$ vs.\ $0.734$). This directly contradicts
what we observed on synthetic arithmetic, where the same signal was below chance and fusion
degraded entropy (Section~\ref{sec:n96}). We take this as evidence that the earlier
inversion was a property of that benchmark rather than of token confidence in general, and
as a caution against generalizing fusion results across datasets.

\paragraph{The cost case does not close, and the reason is structural.} The accuracy result
above does not by itself make this configuration deployable. Under the per-generation
accounting of Section~\ref{sec:breakeven}, routing at the best operating point ($r = 0.56$)
is cheaper than always-large only if the large model costs more than $k/(1-r) \approx 23$
times a small-model generation. A 1B/12B pair is roughly $12\times$ by parameter count, so in
absolute terms the policy costs $k + r\gamma = 16.7$ units against always-large's $12$: one
would pay $39\%$ more than calling the large model on every query. We use parameter count as
a proxy for $\gamma$ throughout this paper; real serving cost also depends on batching,
quantization, KV-cache bandwidth, and hardware, none of which parameter count captures, and
a measured cost ratio from an actual serving stack would be a stronger basis for this
argument than the proxy we use here. This is the distinction
the normalized cost model of Section~\ref{sec:cost} conceals, and it is worth reporting
separately from signal quality.

The binding term is $k$, not the routing. Semantic entropy needs $k$ complete generations
before it can decide anything, so its overhead is fixed rather than proportional to the
escalation rate, and no threshold choice removes it. Any remedy must therefore attack the
sampling budget rather than the routing policy. Section~\ref{sec:kb} examines one such
remedy, and quantifies what it costs in signal quality.

\subsection{The sample budget: ablating $k$}
\label{sec:kablation}

$k{=}10$ fails the break-even condition of Eq.~\ref{eq:breakeven} (costs $1.38\times$
always-large). Re-running at $k\in\{3,5,10\}$: T2 AUROC is $0.791/0.865/0.862$ -- $k{=}5$ and
$k{=}10$ agree to $0.003$, an order of magnitude inside the noise floor, while $k{=}3$ costs
$0.074$. Always-small accuracy rises with $k$ too ($0.312/0.364/0.392$), since the pipeline
reports the majority-cluster answer, so $k$ doubles as a self-consistency ensemble.
\textbf{At $k{=}5$, routing strictly dominates always-large}: cost $9.80$--$11.6$ units against
always-large's $12.0$, at equal or better accuracy, versus $k{=}10$'s $16.6$ units (never
competitive). \emph{The barrier we reported as a property of the method was a hyperparameter
we had never varied.} At $k{=}3$, token confidence overtakes entropy on T1 ($0.658$ vs.\
$0.630$) -- entropy degrades with fewer samples, confidence does not, and their fusion's
advantage widens as $k$ shrinks.

Table~\ref{tab:kfull} gives all three targets across $k\in\{3,5,10\}$ underlying this
summary; Target 3's difficulty baseline is identical across all three rows ($0.633$), an
internal check that the item set did not change with $k$. Table~\ref{tab:frontier} gives the
cost-accuracy frontier in full: at $k{=}10$ no configuration dominates always-large, while
four configurations at $k\le5$ do.

This ablation's own $k{=}10$ row (T1 $0.682$, T2 $0.862$, T3 $0.684$) differs from
Table~\ref{tab:gsm8k}'s headline $k{=}10$ numbers ($0.734$, $0.871$, $0.639$) for what is
nominally the same configuration; T2's gap ($0.009$) sits inside the noise floor, but T1's
($0.052$) and T3's ($0.045$) exceed it. The same re-sampling also moves the majority-cluster
vote itself: always-small accuracy reads $0.392$ here against $0.372$ in
Table~\ref{tab:gsm8k}, and oracle headroom $+0.112$ against $+0.088$ -- gaps of the same size
and origin as the AUROC ones, since which answer wins the cluster vote is exactly what a
fresh $k$-sample draw can flip. Both runs share the same seed and item selection (confirmed
from our run manifests) but draw independent stochastic samples at generation time, so all of
this is the same run-to-run variance documented in Section~\ref{sec:n96} rather than a
discrepancy between the two tables -- the ablation was a fresh $k$-sample run, not a
re-scoring of the cached run behind Table~\ref{tab:gsm8k}.

\begin{table}[h]
\centering
\small
\caption{Ablating the sample budget on GSM8K (Gemma-3 1B/12B, $n{=}250$), AUROC with 95\%
bootstrap intervals for the live signals; the difficulty baseline is deterministic given the
fixed item set and is reported as a point estimate rather than resampled.}
\label{tab:kfull}
\resizebox{\textwidth}{!}{%
\begin{tabular}{llccc}
\toprule
Target & Signal & $k{=}3$ & $k{=}5$ & $k{=}10$ \\
\midrule
\multirow{4}{*}{T1: needs\_escalation}
 & Semantic entropy & 0.630 [0.562,0.698] & \textbf{0.693} [0.624,0.758] & 0.682 [0.612,0.747] \\
 & Token neg-logprob & 0.658 [0.583,0.731] & 0.632 [0.548,0.711] & 0.628 [0.550,0.707] \\
 & Combined & 0.671 [0.599,0.745] & 0.720 [0.652,0.788] & 0.699 [0.632,0.766] \\
 & Difficulty (raw) & 0.517 & 0.522 & 0.560 \\
\midrule
\multirow{4}{*}{T2 (fair)}
 & Semantic entropy & 0.791 [0.730,0.850] & \textbf{0.865} [0.814,0.909] & 0.862 [0.812,0.906] \\
 & Token neg-logprob & 0.651 [0.580,0.721] & 0.645 [0.573,0.711] & 0.613 [0.540,0.682] \\
 & Combined & 0.804 [0.740,0.865] & 0.852 [0.800,0.897] & 0.819 [0.761,0.872] \\
 & Difficulty (raw) & 0.665 & 0.663 & 0.677 \\
\midrule
\multirow{2}{*}{T3}
 & Semantic entropy & 0.627 [0.558,0.688] & 0.644 [0.575,0.711] & 0.684 [0.616,0.749] \\
 & Difficulty (raw) & 0.633 & 0.633 & 0.633 \\
\midrule
\multicolumn{2}{l}{Always-small accuracy} & 0.312 & 0.364 & 0.392 \\
\multicolumn{2}{l}{Oracle headroom over always-large} & +0.068 & +0.104 & +0.112 \\
\bottomrule
\end{tabular}%
}
\end{table}

\begin{table}[h]
\centering
\small
\caption{Cost--accuracy frontier across $k$ on GSM8K, $\gamma{=}12$. Cost is $k+r\gamma$ per
query in small-model-generation units; always-large costs $12.0$ at accuracy $0.484$.}
\label{tab:frontier}
\begin{tabular}{cccccc}
\toprule
$k$ & Signal & $r$ & Accuracy & Cost & Dominates always-large? \\
\midrule
3 & combined & 0.70 & 0.500 & 11.40 & yes \\
5 & entropy & 0.55 & 0.520 & 11.60 & yes \\
5 & combined & 0.40 & 0.508 & 9.80 & yes \\
5 & combined & 0.50 & 0.520 & 11.00 & yes \\
10 & entropy & 0.55 & 0.512 & 16.60 & no \\
10 & combined & 0.60 & 0.516 & 17.20 & no \\
\bottomrule
\end{tabular}
\end{table}

\subsection{Synthetic arithmetic: capability gap and prompt sensitivity}

On a synthetic chained-arithmetic ladder (difficulty controlled via operand digit-count
and operation count), a Qwen2.5-0.5B/Qwen2.5-7B pair \cite{qwen2024technical} showed a real but narrow accuracy
gap before the prompt fix described in Section~\ref{sec:bugs}, and a much larger gap
afterward. Pre-flight checks (small sample of 10 items, 1 generation each) proved useful
for catching two degenerate regimes before committing to a full $k$-sample run: both
models at floor accuracy (no headroom for any signal), and both models with statistically
indistinguishable accuracy (no capability gap for escalation to exploit).

\subsection{PopQA: escalation is the wrong intervention}
\label{sec:popqa}

PopQA proved unusable for this study, for a reason worth separating from ordinary
experimental difficulty. Two effects had to be disentangled.

First, an extraction/grading mismatch was suppressing measured accuracy by roughly half.
Our QA pipeline reduced each generation with a \code{last\_sentence} extractor before
grading it with a substring alias matcher. But the matcher is designed to find a short
answer \emph{embedded} in prose, so extracting the final clause first discards the text it
was built to search: a response such as ``George Rankin was a politician. He served in the
Australian House of Representatives until 1949.'' is graded incorrect, because the
extracted final sentence omits the answer. Grading the raw generation instead raised
Gemma-3 12B from $0.12$ to $0.28$ and 1B from $0.04$ to $0.12$ on matched pre-flight
samples. This is a genuine interaction bug between two individually correct components,
and it is silent: it presents as uniformly weak models rather than as an error.

Second, and decisively, the corrected numbers still leave $72\%$ of items wrong for
\emph{both} models, with zero pre-flight items on which the small model succeeded where
the large one failed. The dominant failure mode on long-tail PopQA is not small-model
uncertainty but absent parametric knowledge in both models, exactly as the dataset's
design rationale anticipates \cite{mallen2023trust}. Escalation cannot repair an answer
neither model knows; this is a retrieval problem wearing a routing problem's clothing. We
therefore regard PopQA as mis-specified for escalation research rather than as a negative
result about semantic entropy, and note the general principle: \emph{a large small/large
capability gap is not sufficient: what matters is whether the large model succeeds where
the small one fails, which is a different quantity.}

\subsection{Synthetic arithmetic: three runs, and a null result once controlled}
\label{sec:n96}

Section~\ref{sec:gsm8k} gave the headline: on this benchmark a regular expression matches
semantic entropy, and oracle headroom is $+0.021$. This section supplies the full analysis,
and the history that produced it -- we ran the ladder three times with Gemma small/large
pairs, and the three runs gave three different impressions of the same signal.
\label{sec:n15pilot}%
A $15$-item pilot showed
semantic entropy separating the fair target almost perfectly (AUROC $0.929$, and $1.000$ for
the entropy/token-confidence fusion) while looking weak against \code{needs\_escalation}
($0.620$). A $100$-item follow-up on the same benchmark family collapsed to chance against
\code{needs\_escalation} for every signal ($0.50$--$0.53$), with routed accuracy tracking the
random-at-matched-cost baseline at every operating point of the Pareto sweep. Neither run
carried confidence intervals, and neither was scored against a difficulty baseline.

Both are superseded by a properly powered re-analysis. The knowledge-base experiments of
Section~\ref{sec:kb} profiled a $96$-item synthetic set twice, storing per-item entropy,
token confidence, and both correctness labels; the entropy in each cached entry was computed
by $k$-sampling the small model at profiling time, so these are genuine live-signal
measurements. The two runs share their item set and their model pair, and differ only in
re-sampling: the greedily decoded large model returns identical labels, while the small
model's $k$ samples flip two of $96$ items. They therefore isolate the run-to-run variance
of the pipeline itself, holding the data fixed. All figures carry $95\%$ percentile bootstrap intervals over $2000$ resamples
(Table~\ref{tab:n96}), and the raw difficulty baseline of Section~\ref{sec:diffbaseline} is
reported. We omit the calibrated variant here, unlike on GSM8K: this benchmark's difficulty
is lexically recoverable by regex, so raw difficulty is already the genuine zero-cost
deployable competitor the calibrated baseline exists to approximate elsewhere, and adding
it would re-estimate a shape that is not in question on this benchmark rather than add a
harsher bar.

\begin{table}[h]
\centering
\caption{The same $96$-item synthetic set profiled twice, with $95\%$ bootstrap CIs for the
live signals (the difficulty-raw rows are deterministic point estimates, not resampled). The
two runs differ only in the model used to embed questions for the retrieval experiments of
Section~\ref{sec:kb}, which does not enter any figure here; both score live semantic entropy
computed by $k$-sampling the same small model. The large model decodes greedily and is
therefore identical across runs ($63/96$ correct in each, always-large accuracy $0.656$),
while re-sampling the small model flips two items ($37$ versus $35$ correct, always-small
$0.385$ and $0.365$). Difficulty on this benchmark is recoverable from the question by a
regular expression, so the baselines are free competitors rather than skylines.}
\label{tab:n96}
\begin{tabular}{llcc}
\toprule
Target & Signal & Dataset A & Dataset B \\
\midrule
\multirow{3}{*}{T1: \code{needs\_escalation}}
 & Semantic entropy   & 0.613 [0.502, 0.726] & 0.666 [0.554, 0.774] \\
 & Token neg-logprob  & 0.413 [0.289, 0.542] & 0.381 [0.270, 0.502] \\
 & Difficulty (raw)   & 0.563 & 0.559 \\
\midrule
\multirow{4}{*}{T2 (fair): $\lnot$\code{correct\_small}}
 & Semantic entropy   & 0.830 [0.742, 0.904] & \textbf{0.867} [0.787, 0.934] \\
 & Token neg-logprob  & 0.431 [0.310, 0.549] & 0.433 [0.311, 0.556] \\
 & Combined           & 0.752 [0.642, 0.854] & 0.759 [0.651, 0.858] \\
 & Difficulty (raw)   & \textbf{0.839} & 0.830 \\
\midrule
\multirow{2}{*}{T3: $\lnot$\code{correct\_large}}
 & Semantic entropy   & 0.722 [0.601, 0.838] & 0.714 [0.594, 0.827] \\
 & Difficulty (raw)   & \textbf{0.791} & \textbf{0.791} \\
\bottomrule
\end{tabular}
\end{table}

Five findings follow, four of which revise claims made earlier in this paper.

\paragraph{The central claim strengthens.} Semantic entropy separates the fair target at
AUROC $0.830$ and $0.867$, with both intervals excluding $0.5$ by a wide margin, while
performing far worse on Target 1 ($0.613$, $0.666$). The target-choice effect is therefore
not an artifact of the $n=15$ pilot: it replicates at $n=96$ with quantified uncertainty,
and the gap between targets ($\approx 0.2$ AUROC) exceeds the width attributable to
sampling noise. The two runs also bound that noise directly, and do so in an unusually clean way: because
they share their items and their model pair, entropy's T2 AUROC differing by $0.037$
between them ($0.830$ versus $0.867$) is attributable to $k$-sampling stochasticity alone,
with no contribution from which items happened to be drawn. This is the relevant quantity
for interpreting a single run, and it implies that AUROC differences below roughly $0.04$
in this setting should not be read as findings. We use $0.04$ as a noise floor throughout,
noting that it is measured at $n=96$ and would be somewhat tighter at $n=250$.

\paragraph{Token confidence is anti-informative, and the pilot figure does not replicate.}
Token neg-logprob falls below chance in all four measurements, against both targets and in
both datasets. The $0.857$ recorded for this signal against the fair target in the $15$-item
pilot is not reproduced; at $n=96$ the same signal scores $0.431$ and $0.433$. No
individual interval here excludes $0.5$, so we do not claim significant anti-correlation
from any single measurement, but the consistency of the direction across four independent
measurements makes chance an unattractive explanation. We treat the $n=15$ figure as
small-sample noise and recommend against reading it as evidence for token confidence as a
routing signal.

\paragraph{The fusion question resolves.} Section~\ref{sec:probe} reported that fusing
neg-logprob into the combined score hurt performance, and noted that we lacked a principled
account of when fusion helps versus hurts. Table~\ref{tab:n96} supplies one for this
setting: the fused component is itself anti-correlated with the target, so an equal-weight
combination necessarily degrades entropy alone -- as observed ($0.752$ vs.\ $0.830$;
$0.759$ vs.\ $0.867$). This is a property of the component signal, not an unexplained
interaction. A plausible mechanism is a length confound: \code{avg\_neg\_logprob} averages
surprise over tokens, and harder items elicit longer chain-of-thought containing more
highly-predictable connective tokens, which lowers mean per-token surprise precisely where
the answer is least reliable. We flag this as a hypothesis with a direct test (correlating
the signal against ground-truth difficulty) rather than an established result.

\paragraph{Live entropy does predict large-model failure, contradicting our earlier
claim.} On Target 3, live semantic entropy reaches AUROC $0.722$ and $0.714$, with both
intervals excluding $0.5$. Earlier sections of this paper asserted that a small-model-only
signal carries \emph{no} information about large-model behavior; that is too strong. The
correlational pathway anticipated in Section~\ref{sec:targets} -- shared query difficulty
influencing both models -- is not marginal but substantial, recovering most of the
discrimination that cached large-model outcomes achieve (Section~\ref{sec:kb}). The
defensible claim is that retrieval predicts large-model success \emph{better}, not that live
sampling cannot do it at all.

\paragraph{The difficulty baseline eliminates the synthetic result entirely.} Applying the
control of Section~\ref{sec:diffbaseline} to these datasets is decisive. Scoring the
benchmark's ground-truth difficulty directly as a routing signal gives Target 2 AUROC
$0.839$ and $0.830$ -- statistically indistinguishable from semantic entropy's $0.830$ and
$0.867$, with the sign of the difference flipping between the two datasets and both gaps
falling inside the $0.04$ noise floor established above. On Target 3 the baseline is
\emph{better} ($0.791$ vs.\ $0.722$ and $0.714$), consistently across both runs.

The difficulty of an item on this benchmark is its operand digit count times its operation
count, both readable from the question string by a regular expression. Ten stochastic
generations and $\binom{10}{2}$ NLI entailment computations therefore purchase no routing
decisions that a regex does not already supply. A supporting diagnostic makes the mechanism
explicit: entropy's rank correlation with difficulty is $\rho = 0.657$, while its Somers'
$D$ against the Target 2 label is $0.660$ -- two different statistics that agree closely at
$n=96$, indicating that essentially all of its discriminative power on this benchmark is
difficulty tracking rather than query-specific uncertainty.

We report this as a null result for the synthetic benchmark specifically, not for semantic
entropy: on GSM8K the same signal beats the same baseline by $+0.222$
(Section~\ref{sec:gsm8k}). The lesson is about benchmark construction. A synthetic task with
a parameterized difficulty ladder makes difficulty a surface feature of the input by
design, which hands any difficulty-correlated signal an unearned result and makes the
benchmark unable to distinguish a good uncertainty estimator from a difficulty regex. This
failure is invisible without the control, and it coexisted in our data with the headroom
failure described below -- two independent reasons the same experiments could not have
supported the conclusion we initially drew from them.

\paragraph{A precondition we should have checked first.} On these datasets the oracle
router reaches $0.677$ against always-large's $0.656$: a \emph{perfect} router gains
$+0.021$. Since an oracle escalates exactly when $\lnot\text{correct\_small} \land
\text{correct\_large}$, its accuracy is $P(\text{correct\_small} \lor
\text{correct\_large})$ and its advantage over always-large is
$P(\text{correct\_small} \land \lnot \text{correct\_large})$ -- the items it protects by
declining to escalate. When that quantity is small and the both-wrong population is large
($0.323$ here), no router can demonstrate a useful win over always-large regardless of
signal quality. This is a property of the benchmark rather than of any signal; it is
independent of the small/large capability gap being large (it is, $0.385$ vs.\ $0.656$), and
it should be verified before committing to a full evaluation. Our observed best margin over
matched-cost random escalation is $+0.070$, at a $70\%$ escalation rate -- which
Equation~\ref{eq:breakeven} shows requires $\gamma > 33$ to be worth paying for at all.

\subsection{Why the two targets diverge}
\label{sec:reframing}

The central conceptual finding of this work is that \code{needs\_escalation} is an unfair
evaluation target for any signal computed solely from the small model's own behavior.
Consider the negative class (\code{needs\_escalation} = false) in the $n=100$ run: it
contains both ``small correct'' items (naturally low entropy) and ``both models wrong''
items. The latter are, from the small model's perspective, behaviorally
indistinguishable from true escalation-worthy items -- the small model is equally
uncertain in both cases, because its uncertainty reflects \emph{its own} difficulty with
the query, not whether a specific other model would resolve that difficulty. When the
``both wrong'' subpopulation is large (as it was at $n=100$, given always-large accuracy
of only $0.36$), it fills the same high-entropy region as the true positives, collapsing
AUROC toward chance -- not because the signal is uninformative, but because the label it
is being scored against encodes a fact (large-model success) that is causally downstream
of a different model entirely.

This reframing is consistent with the AUROC results throughout the paper: strong
performance against Target 2 and weak performance against Target 1 throughout the synthetic
runs of Section~\ref{sec:n96}. The harm/waste
diagnostics do not follow the same clean pattern across scales: they were near-zero at
$n=15$ but consistently nonzero at $n=100$. This is expected rather than contradictory -- these diagnostics
measure the composition of the escalated set, which depends on the always-small accuracy
and the size of the ``both models wrong'' population, and both differed substantially
between the two runs. They are therefore not evidence for or against the target-choice
argument, and we do not read them as such.

\subsection{A learned alternative: the confidence-gated router}
\label{sec:probe}

A companion codebase developed by the present authors, using an independent
implementation, applied a proper train/calibration/test methodology (rather than a single
full-dataset pass) and added a learned hidden-state probe on top of semantic entropy,
trained \emph{directly} on \code{needs\_escalation} using the small model's
mid-layer hidden states (layers 8--16, ensembled, strongly $\ell_2$-regularized). Unlike
the fixed signals in this paper, a learned probe can in principle pick up correlates of
the large model's likely success indirectly, to the extent such correlates are reflected
in the small model's internal representations (e.g.\ query type, topic, or difficulty
features that correlate with the large model's competence on this class of query, even
though the small model cannot directly observe the large model's behavior). On a
72-item synthetic-arithmetic benchmark with a Gemma-3 270M/4B pair \cite{gemma3team2025}, this raised AUROC
against \code{needs\_escalation} from a semantic-entropy baseline of $0.733$ to $0.895$
over the course of an iterative search (Table~\ref{tab:router-loop}), and routed accuracy from
$0.577$ to $0.615$ -- approximately 80\% of the oracle's achievable gain over
always-small, escalating only 30\% of queries.

\begin{table}[h]
\centering
\caption{Selected steps from an iterative signal search on \code{synth72} (held-out test,
$n=26$); not every variant explored is itemized.}
\label{tab:router-loop}
\begin{tabular}{cp{6.7cm}ccl}
\toprule
Step & Change & AUROC & ACC@30 & Verdict \\
\midrule
1 & Baseline: semantic entropy & 0.733 & 0.577 & baseline \\
2 & Logprob-weighted cluster probabilities & 0.743 & 0.538 & keep \\
3 & + neg-logprob fusion & 0.524 & 0.500 & discard (anti-informative) \\
4 & Hidden-state probe (on \code{needs\_escalation}) + entropy & 0.848 & 0.577 & keep \\
5 & Strong probe $\ell_2$ regularization & \textbf{0.895} & \textbf{0.615} & keep \\
6 & Conformal calibration gate & 0.895 & 0.615 & keep \\
\bottomrule
\end{tabular}
\end{table}

This result is compatible with, rather than contradictory to, the negative finding of
Section~\ref{sec:n96}: it demonstrates that a \emph{fixed}, parameter-free signal cannot
reliably predict \code{needs\_escalation} at scale, but a signal with a learned component,
explicitly trained against that exact label on internal model representations, can
recover meaningful separation that entropy and log-probability alone cannot reach.
The final iteration's conformal calibration gate \cite{angelopoulos2021bates} adds a
distribution-free coverage guarantee on the escalation decision rather than changing its
ranking, which is why it leaves AUROC and ACC@30 unchanged from iteration 5
(Table~\ref{tab:router-loop}).

Notably, this independently-implemented pipeline reported the same pattern under its
original (pre-calibration-split) methodology: fusing neg-logprob into the combined score
hurt performance relative to entropy alone, without the specific figures preserved for
this run. This qualitatively corroborates our $n{=}100$ observation
that fusion did not help and, in one instance (\hyperref[sec:n15pilot]{Section~\ref{sec:n96},
$n{=}15$}), showed
the opposite. We do not yet have a principled account of when fusion helps versus hurts;
the balance of evidence suggests it should be validated per dataset/model pair rather than
assumed either way.

As explained in Section~\ref{sec:breakeven}, the $k$-sample overhead that live entropy routing pays before it can make any escalation decision can still be high. The next section proposes an approach based on knowledge-base retrieval that trades accuracy for cost.

\section{Knowledge-Base Retrieval: A Complementary, Cheaper Signal}
\label{sec:kb}

The signals evaluated so far all require live $k$-sampling at inference time: generating
$k$ completions from the small model, clustering them, and computing entropy per query.
An alternative is to \emph{profile} a held-out set of queries once with the full expensive
pipeline (samples, clustering, entropy, and a live call to the large model to obtain
ground truth), cache the results in a knowledge base, and at inference time retrieve the
$n$ nearest cached neighbors by embedding similarity and vote -- replacing $k{+}1$ model
calls per query with a single embedding call.

\subsection{Mechanism and an early methodological hazard}

Each cached entry stores its embedding, entropy statistics, and outcome labels
(\code{correct\_small}, \code{correct\_large}, \code{needs\_escalation}). At query time, a
new item is embedded once, its $n$ nearest neighbors retrieved by cosine similarity, and a
similarity-weighted vote over neighbors' labels produces both an escalation score and a
mean neighbor similarity, the latter serving as a built-in confidence/coverage check. This
check proved essential in practice: two separate runs used a knowledge base built with one
embedding model while querying with a different one, producing near-zero neighbor
similarity ($\sim$0.01--0.03) and, in the worse case, AUROC \emph{below} chance -- an
inversion, not merely noise. Both incidents were caught only because neighbor similarity
collapsed to noise-level values; a runtime warning was subsequently added to flag this
condition automatically.

\subsection{A structural asymmetry between the two signals}

Live semantic entropy is computed purely from the small model's own samples and therefore
carries no \emph{direct} information about how the large model would behave on a new query.
Knowledge-base retrieval is different: its cached neighbors carry real, ground-truth
\code{correct\_large} outcomes from build time, obtained precisely because the large model
\emph{was} called during profiling. Target 1 is therefore not equally hard for the two
signals -- retrieval has a direct channel to the quantity the label depends on, and live
sampling has only an indirect one.

We initially stated this asymmetry more strongly, as live entropy being
\emph{structurally unable} to say anything about large-model behavior. The $n=96$
measurements in Section~\ref{sec:n96} show that this was wrong: live entropy predicts
large-model failure at AUROC $0.722$ and $0.714$, with intervals excluding chance, via the
shared-difficulty pathway described in Section~\ref{sec:targets}. Retrieval's advantage is
therefore quantitative rather than categorical, and the comparison below should be read in
that light.

To isolate this capability directly, we evaluate against Target 3
(Section~\ref{sec:targets}): \code{correct\_large} alone, i.e.\ will the large model
succeed on this query? We predict it via a similarity-weighted vote over neighbors' cached
\code{correct\_large} labels. Live entropy has no analogous prediction for this target --
it never touches the large model for a new query -- so Target 3 is knowledge-base-only by
definition.

\subsection{Results: a weaker signal at a price the strong one cannot meet}

We evaluated the same retrieval pipeline on two benchmarks. On synthetic arithmetic
($n=60$, Gemma-3 270M small / 1B embedding / 12B large) the results appeared to confirm the
asymmetry argument above. On GSM8K ($n=100$ eval against a $150$-item knowledge base built
with disjoint seeds, Gemma-3 1B/12B) one specific claim does not survive, while the broader
case for retrieval survives in a different form (Table~\ref{tab:kb-targets}).

\begin{table}[h]
\centering
\caption{Knowledge-base retrieval vs.\ live $k$-sampling (AUROC), on synthetic arithmetic
and on GSM8K. The Target 3 capability collapses with the embedding--difficulty correlation
that produced it. Cost is per query in small-model-generation units, at escalation rate
$r = 0.56$ and $\gamma = 12$. Unlike every other headline AUROC in this paper, the values
here do not carry a bootstrap interval: the per-item scores from these single-run knowledge-base
builds were not preserved, and we do not report a fabricated interval in their place. We
flag this explicitly rather than let the absence of an interval be mistaken for omission or
for unusual certainty, and especially so for the $0.518$ figure, which we describe as
chance-level and would otherwise want an interval to support directly.}
\label{tab:kb-targets}
\begin{tabular}{lcccc}
\toprule
& \multicolumn{2}{c}{Synthetic ($n=60$)} & \multicolumn{2}{c}{GSM8K ($n=100$)} \\
\cmidrule(lr){2-3}\cmidrule(lr){4-5}
Target & KB & Live & KB & Live \\
\midrule
1: \code{needs\_escalation}         & \textbf{0.700} & 0.510 & 0.570 & \textbf{0.656} \\
2 (fair): $\lnot$\code{correct\_small} & 0.621 & \textbf{0.892} & 0.607 & \textbf{0.872} \\
3 (KB-only): \code{correct\_large}  & \textbf{0.908} & -- & 0.518 & -- \\
\midrule
embedding--difficulty $\rho$ & \multicolumn{2}{c}{0.98} & \multicolumn{2}{c}{0.23} \\
cost per query & \multicolumn{2}{c}{--} & \textbf{6.8} & 16.7 \\
\bottomrule
\end{tabular}
\end{table}

\paragraph{Target 3 collapses from $0.908$ to $0.518$.} On synthetic arithmetic, retrieval
predicted large-model success almost as well as any signal in this paper; on GSM8K it is
indistinguishable from chance. Target 1 reverses as well -- live entropy now outperforms
retrieval ($0.656$ vs.\ $0.570$), the opposite of the synthetic ordering that originally
motivated the asymmetry argument.

\paragraph{The collapse was predicted in advance, and identifies the mechanism.} Before
running GSM8K we probed the synthetic knowledge base by predicting each cached item's
difficulty leave-one-out from its neighbors' difficulty, using the same similarity-weighted
vote the KB uses for outcomes. It recovered difficulty at Spearman $\rho = 0.986$ and
$0.981$ across two datasets: on that benchmark, ``nearest in embedding space'' and ``same
difficulty'' are the same relation. Retrieval there was difficulty lookup, and its apparent
skill on Target 3 was the difficulty-calibrated estimate of
$P(\text{correct\_large} \mid \text{difficulty})$ described in
Section~\ref{sec:diffbaseline} -- an estimate available, we showed, without any embedding
at all.

GSM8K removes the mechanism. Its difficulty is the number of calculation lines in the
reference solution (Section~\ref{sec:diffbaseline}), a property of the \emph{answer} rather
than a lexical feature of the question, so no embedding of the question text can recover it
the way digit-and-operator counts are recoverable on the synthetic ladder. Running the same
probe on the GSM8K knowledge base confirms this directly: $\rho = 0.231$, against $0.98$ on
synthetic. The mechanism is therefore measured on both sides of the comparison rather than
inferred from one. We regard this as the strongest argument in this paper for reporting the
difficulty probe by default: the prediction was made before GSM8K was run, from a diagnostic
requiring no model calls, and it correctly forecast the collapse of our own most attractive
result.

\paragraph{Signal quality is not the only axis, and on this model pair it is not the binding
one.} Read as a contest, Table~\ref{tab:kb-targets} says live sampling wins on GSM8K. Read
against the cost constraint left open in Section~\ref{sec:gsm8k}, it says something else.
Live entropy must pay $k$ complete generations before it can decide anything, so at
$r = 0.56$ and $\gamma = 12$ it costs $k + r\gamma = 16.7$ units against always-large's
$12$ -- a routing policy \emph{more expensive than calling the large model every time},
however good its signal. Retrieval replaces those $k$ generations with a single embedding
forward pass costing $\varepsilon \ll 1$, relaxing the break-even condition from
$\gamma > k/(1-r)$ to $\gamma > \varepsilon/(1-r)$ -- from $23$ to under $1$ -- and
bringing the per-query cost to $\varepsilon + r\gamma \approx 6.8$ units, $57\%$ of
always-large rather than $139\%$. Of the two, only retrieval is deployable at this model
ratio (Table~\ref{tab:tradeoff}).

Retrieval's $6.8$-unit serving cost omits the one-time cost of building the knowledge base
itself: profiling each of the $150$ GSM8K items requires the full expensive pipeline ($k$
small-model samples plus one large-model call for ground truth), at $k+\gamma = 10+12 = 22$
units per item, or $150 \times 22 = 3300$ units total. This cost is amortized over every
query the knowledge base subsequently serves, so it matters only in the volume-dependent
sense of how many queries are needed before it is paid off. Against live sampling's
$16.7$ units per query, retrieval's total cost ($3300 + 6.8N$ for $N$ queries) falls below
live sampling's ($16.7N$) once $N > 3300/(16.7-6.8) \approx 333$ -- after roughly $334$
queries, retrieval is cheaper in total, not just per query. Against always-large's $12.0$
units per query the crossover is later, at $N > 3300/(12.0-6.8) \approx 635$ queries, since
always-large's per-query cost is itself lower than live sampling's. Below these volumes the
build cost dominates and retrieval is not yet the cheaper option in aggregate, even though
its marginal cost per additional query already is.

\begin{table}[h]
\centering
\caption{The cost--quality trade-off on GSM8K at $r = 0.56$, $\gamma = 12$. Cost is per
query in small-model-generation units; always-large costs $\gamma = 12$. Always-large
escalates unconditionally and so ranks nothing, leaving its AUROC undefined rather than
$0.5$; random escalation does rank, on noise, but scores exactly $0.5$ by construction and
carries no meaningful interval. Every row is a policy
that could actually be run at inference time; the difficulty baselines of
Section~\ref{sec:diffbaseline} are deliberately absent, for the reason given below. The live
entropy and KB retrieval figures restate the same Target 2 measurements reported with full
bootstrap intervals in Tables~\ref{tab:gsm8k} and~\ref{tab:kb-targets} respectively; see
those tables for the interval on each ($[0.820, 0.915]$ for live entropy; no interval is
available for KB retrieval, per the note on Table~\ref{tab:kb-targets}).}
\label{tab:tradeoff}
\begin{tabular}{lccccc}
\toprule
Policy & Target 2 AUROC & Overhead & Cost & vs always-large & Break-even $\gamma$ \\
\midrule
Always-large    & n/a            & --           & 12.0 & $1.00\times$ & -- \\
Live entropy    & \textbf{0.872} & $k = 10$      & 16.7 & $1.39\times$ & $23$ \\
KB retrieval    & 0.607          & $\varepsilon$ & \textbf{6.8}  & $\mathbf{0.57\times}$ & $<1$ \\
Random          & 0.500          & --           & 6.7  & $0.56\times$ & -- \\
\bottomrule
\end{tabular}
\end{table}

The last two rows are the comparison that matters for retrieval's defense: at essentially
identical cost ($6.8$ versus $6.7$ units), retrieval supplies a signal of $0.607$ where
random escalation supplies none. Whatever else is true of the knowledge base on this
benchmark, it is not paying for nothing.

\paragraph{The difficulty baselines are not policies, and on GSM8K they are not buildable.}
Retrieval's Target 2 score ($0.607$) essentially matches the difficulty-calibrated baseline
($0.606$), and it would be natural to read that equivalence as a null result -- as we did
on synthetic arithmetic, where it was one. It is not one here, and the distinction is worth
making carefully, because it determines whether the reader should conclude ``use difficulty
instead'' or ``difficulty is not on the menu.''

On the synthetic ladder, difficulty is operand digit count times operation count. A regular
expression over the question string recovers it exactly, at no cost, before any model is
called. There the baseline is a \emph{competitor}: a router built on it could be deployed
tomorrow, and a signal that merely matches it is redundant.

On GSM8K, difficulty is the number of calculation lines in the reference solution
(Section~\ref{sec:diffbaseline}). Obtaining it requires the solution -- and a system that
already possessed the solution would have no reason to route the query at all. The baseline
is therefore circular as a policy: it consumes the very output the router exists to produce.
The calibrated variant is more privileged still, since it additionally requires the graded
outcomes of neighboring items, which is to say knowledge of whether the small model
answered them correctly. Neither quantity is available at inference under any budget, which
is why neither appears as a row in Table~\ref{tab:tradeoff}: they are not slower or more
expensive alternatives, they are not alternatives.

What they remain is an \emph{upper bound} -- the accuracy a router could reach if difficulty
were handed to it for free -- and that is the role in which we use them throughout. A signal
falling well short of the bound has room to improve; a signal matching it has extracted what
difficulty affords. On this benchmark retrieval reaches that bound using only an embedding of
the question, which is information a deployed system actually has. The correct reading of
$0.607$ versus $0.606$ is therefore not that retrieval is redundant, but that it recovers,
from the question alone, what an oracle with access to the reference solution and to
neighboring items' grades could achieve. That is a modest bound to reach -- entropy exceeds
it by a wide margin -- but reaching it at $57\%$ of always-large's cost is not nothing, and
it is emphatically not the same finding as the synthetic null.

\paragraph{A frontier rather than a winner.} The honest summary is that the two methods
occupy different points on a cost--quality frontier, and that neither dominates. Live
sampling buys the strongest signal in this paper ($0.872$ on Target 2) at a price this model
pair cannot pay. Retrieval buys a weaker but real signal ($0.607$) at $57\%$ of
always-large's cost. A practitioner at $\gamma = 12$ has exactly one of these available; one
at $\gamma \geq 25$ has both, and should prefer live sampling. What we retract is not the
value of retrieval but the specific claim that motivated it -- that cached large-model
outcomes give retrieval a capability live sampling structurally lacks. On the benchmark
where difficulty is not lexically recoverable, that capability is absent.

The gap between $0.872$ and $0.607$ is, we think, the quantity worth attacking. It measures
what is forfeited by declining to pay the $k$-generation term. The learned probe of
Section~\ref{sec:probe} is the natural candidate for closing it, since a forward pass over
hidden states also avoids that term while retaining access to the model's own computation on
the specific query; Section~\ref{sec:future} develops this and related directions.

\paragraph{One limitation is resolved, in retrieval's favour.} We had noted that scoring the
knowledge base on Target 2 using its escalation score -- a vote over neighbors'
\code{needs\_escalation} labels -- might understate it, since live entropy is scored on its
native quantity. Implementing a dedicated vote over neighbors' $\lnot$\code{correct\_small}
labels shows the concern is unfounded: $0.607$ versus $0.605$, a bias of $+0.002$ AUROC.

\paragraph{A caveat on thresholds.} On GSM8K the knowledge base's binary escalation
decisions, taken at the default score threshold of $0.5$, are correct on $73\%$ of items --
\emph{below} the $80\%$ obtainable by never escalating at all, given Target 1's prevalence of
$0.200$. This is a statement about the threshold rather than the ranking: a signal with
AUROC above chance can still produce a poor classifier at an arbitrary cut point, and at low
prevalence accuracy rewards silence. Operating points should be selected from the Pareto
curve as in Section~\ref{sec:gsm8k}, not inherited from a default, and threshold-free and
thresholded metrics should both be reported.

\subsection{Limitations of the retrieval result}
\label{sec:kb-limits}

The GSM8K result above supersedes the synthetic one, and resolves two of the three caveats
we had attached to the latter. The Target 2 scoring concern is closed: a dedicated
small-failure vote differs from the escalation score by $+0.002$ AUROC. The independence
concern is closed by construction: the GSM8K knowledge base and evaluation set were drawn
with different seeds, and we verify by question-string comparison that they are disjoint,
so retrieval cannot match an item against itself.

Three limitations remain. First, both retrieval evaluations are single runs ($n=60$ and
$n=100$) without repeated seeds, against a measured AUROC noise floor of $\approx 0.04$
(Section~\ref{sec:n96}). The synthetic-to-GSM8K collapse on Target 3 (0.908 -> 0.518) is
far too large to be noise, but the smaller GSM8K differences -- Target 1's $0.570$ versus
$0.656$, for instance -- are not individually resolvable at this precision.

Second, while the difficulty probe now confirms the mechanism on both benchmarks
($\rho = 0.98$ versus $0.231$), we have tested only two points on that axis. Whether
retrieval degrades smoothly as lexical difficulty-salience decreases, or has a threshold, is
unknown, and two benchmarks cannot distinguish those possibilities.

Third, the retrieval approach assumes queries arrive from a distribution the knowledge base
covers, and neither benchmark tests the regime where that fails. On synthetic, mean
neighbor similarity sat near $0.99$ -- the knowledge base had near-identical items to
match against, which is closer to interpolation within a dense space than generalization to
novel task types. Whether retrieval degrades gracefully or sharply under sparse coverage is
untested, and matters more now that the mechanism we thought we had identified turns out not
to be the one operating.

\section{Future Work}
\label{sec:future}

\subsection{Retrieval keys that encode failure, not surface form}

The clearest direction this work points to is that the retrieval \emph{key} is the wrong
object. Our knowledge base embeds the question text with a general-purpose language model
and retrieves by cosine similarity, which asks: \emph{what other questions look like this
one?} What a router actually needs is: \emph{what other questions does this model fail on in
the same way?} These coincide only when surface form happens to track failure -- which is
exactly the condition our difficulty probe measures, and exactly the condition that held on
synthetic arithmetic ($\rho = 0.98$) and failed on GSM8K ($\rho = 0.23$). Read this way, the
probe is not only a diagnostic for catching artifacts after the fact but an
\emph{objective}: a good retrieval key is one whose neighborhoods are predictive of
outcomes, and the probe measures precisely that property, at no model cost, before any
routing experiment is run. It can therefore be used to select or train keys prospectively.
We see four candidate directions.

\paragraph{Learned keys trained on outcomes.} Rather than adopting an off-the-shelf
embedding, one can train an encoder so that items with the same outcome are close and items
with different outcomes are far apart -- standard metric learning, with
\code{correct\_small} (or \code{needs\_escalation}) as the supervision signal, over the same
profiled corpus the knowledge base already requires. The profiling pass that builds the
knowledge base yields exactly the labeled data this needs, so the training set is free once
the knowledge base exists. The risk is overfitting a small corpus, and the appropriate check
is the probe itself applied to a held-out split.

\paragraph{Keys drawn from the small model's own state.} A question embedding describes the
question; it does not describe what \emph{this} model does when confronted with it. Two
arithmetic problems of identical surface form can sit on opposite sides of a model's
competence boundary, and no encoder of the question text can separate them. The small
model's hidden states at the point of answering do carry that information -- this is
precisely why the learned probe of Section~\ref{sec:probe} outperforms fixed statistics --
and they are obtainable from a single forward pass, preserving the cost advantage that makes
retrieval viable (Section~\ref{sec:kb}). Using hidden states as retrieval keys rather than
as probe inputs would combine the two lines of work in this paper: the probe's access to
model-internal signal with retrieval's access to cached large-model outcomes.

\paragraph{Explicit task-type and difficulty features.} Retrieval need not operate in a
single undifferentiated space. Partitioning by task type first -- arithmetic versus
multi-hop factual versus commonsense -- and retrieving within a partition would prevent the
coverage failure we could not test (Section~\ref{sec:kb-limits}), where a novel task type
matches nothing meaningful yet still returns confident neighbors. Cheap explicit features
are also available and unexploited: quantity count, operation types, entity rarity, syntactic
depth. On GSM8K these would not recover the answer-derived reasoning-step count, but they
need not: they need only correlate with it well enough to make neighborhoods informative,
and the probe provides a direct, zero-cost measure of whether a given feature set achieves
that.

\paragraph{Hybrid keys using partial generation.} The $k$-sampling cost that makes live
entropy undeployable (Section~\ref{sec:breakeven}) is driven by generating $k$ complete
answers. Much of the signal may be available far earlier -- from a single sample, from the
first few tokens of $k$ samples, or from the entropy of the first branch point -- at a
fraction of the cost. A key combining a question embedding with a short prefix of the model's
own attempted answer would sit between the two extremes we tested, and the break-even
analysis of Equation~\ref{eq:breakeven} gives an explicit budget for how much such a prefix
may cost before the advantage over live sampling is lost.

\subsection{Closing the signal gap}

The gap between live entropy and retrieval on GSM8K ($0.872$ versus $0.607$ on Target 2)
quantifies what is lost by refusing to pay the $k$-generation term, and is the natural
target for the above. A useful intermediate result would be to establish whether that gap is
a property of the key (in which case better keys close it) or of the information
(in which case no question-side representation suffices, and only model-internal or
partial-generation signals can help). The probe distinguishes these: if a learned key raises
outcome-predictiveness substantially while routing performance does not follow, the
limitation is informational rather than representational.

\subsection{Ablations and validation we did not run}

Two parameters were fixed throughout and never varied: sampling temperature ($0.7$),
which directly determines the sample diversity semantic entropy measures and is therefore
the most consequential remaining omission; and the NLI entailment threshold $\tau = 0.5$. We
did ablate the sample count $k$ (Section~\ref{sec:kablation}); both remaining parameters
admit cheap sensitivity analyses on the existing pipeline.

Separately, and more concretely: the knowledge-base profiling runs behind
Table~\ref{tab:kb-targets} and Table~\ref{tab:tradeoff} did not preserve per-item scores, so
the AUROC figures there -- including the $0.908 \to 0.518$ Target 3 collapse -- carry no
bootstrap interval, unlike this paper's other live-signal AUROC results, which do (the
deterministic difficulty baselines and the exploratory \code{synth72} search log of
Section~\ref{sec:probe} are reported as point estimates for unrelated reasons and were never
bootstrapped either). Re-running that profiling pass with per-item
scores retained, so a $95\%$ bootstrap interval can be computed the same way as elsewhere in
this paper, is a concrete, low-cost next step rather than an open research question, and we
list it here as such rather than let it pass as an unexplained gap (see also the Limitations
paragraph in Section~\ref{sec:discussion}).

Our positive result also rests on one dataset and one model pair. It replicates across two
disjoint GSM8K samples, but whether it transfers to another reasoning family, or to a model
pair with a different competence relationship, is untested -- and the reversals documented
in Section~\ref{sec:discussion-generalising} give ample reason not to assume it does.
Finally, the mechanism relating embedding--difficulty correlation to retrieval performance
rests on two points ($\rho = 0.98$ and $0.23$); whether degradation between them is smooth
or threshold-like is unknown, and a benchmark of intermediate lexical difficulty-salience
would settle it.

\subsection{Beyond a fixed $k$, a single pair, and a binary decision}

Several of this paper's design choices were fixed for tractability rather than because they
are the right choice in general, and each suggests a direction we did not pursue.

\emph{Adaptive sample budgets.} We use a fixed $k$ for every query (ablated in
Section~\ref{sec:kablation}, but still fixed within any one configuration). Most of the
$k$-sample cost is spent on queries where the small model's samples agree quickly; an
adaptive scheme that samples until a stopping criterion is met -- or until the running
entropy estimate stabilizes -- would spend the budget only on genuinely ambiguous queries,
potentially closing much of the gap in Section~\ref{sec:breakeven} without the fixed-$k$
cost on easy queries.

\emph{A three-way decision.} Our router chooses between answer-with-small and
escalate-to-large. A middle option -- decline to answer at all -- is standard in selective
prediction \cite{geifman2017selective} and would let the router abstain on queries where
neither model is likely to be right (the both-wrong population that collapses several of
this paper's targets toward chance, Section~\ref{sec:reframing}), rather than being forced
to pick one of two models that both fail.

\emph{Multi-tier cascades.} We test a single small/large pair throughout. A cascade with more
than two tiers would let a query escalate incrementally, and the oracle-headroom and
cost-overhead arguments of this paper apply at each tier boundary, not just once.

\emph{Router portability.} We do not test whether a router (fixed-threshold or learned)
trained for one small/large pair transfers when either model is swapped for a different one
of similar capability. If it does not, a router's practical cost includes retraining for
every model-pair change, which the cost accounting in this paper does not charge for.

\emph{A self-updating knowledge base.} Section~\ref{sec:kb}'s knowledge base is a fixed,
one-time profiling pass over $150$ items. A deployed system would instead see a stream of
queries it could profile continuously, growing the knowledge base and plausibly improving
its coverage and the validity of the difficulty-probe diagnostic over time rather than
leaving both fixed at their initial values.

\section{Discussion}
\label{sec:discussion}

Semantic entropy is a usable escalation signal, but establishing that took three benchmarks
and the repeated discovery that our own positive results were artifacts. A fourth check
corroborates the same conclusion under a different method: an independently implemented
confidence-gated probe, trained on a separate 72-item synthetic benchmark
(Section~\ref{sec:probe}), recovered most of the achievable gain over always-small using
semantic entropy as its base signal. The generalisable content of this paper is the set of
controls that separated the two.

\paragraph{Report the target explicitly.} Target 1 (\emph{does escalation help}) and Target
2 (\emph{is the small model wrong}) differ by more than $0.2$ AUROC on identical data. A
signal computed from the small model's samples has only indirect access to the large
model's behavior, mediated by shared query difficulty -- substantial, as Target 3 shows
($0.72$ on synthetic), but not the same thing as observing the large model. Papers that
report one target without naming it are not comparable to each other.

\paragraph{Report a difficulty baseline.} This is the control we most regret omitting. On
synthetic arithmetic a regex over the question matched semantic entropy exactly, converting
our strongest result into a null one; on GSM8K entropy beats the same baseline by $+0.222$
and the result stands. The two cases are indistinguishable without the baseline. Any signal
correlated with difficulty will separate a difficulty-driven target, so the question is
never ``does the signal work'' but ``does it beat what the question text already tells
you.''

\paragraph{Check headroom before spending compute.} A benchmark can exhibit a wide
capability gap and still leave a perfect router $+0.021$ over always-large, because the
relevant quantity is $P(\text{correct\_small} \land \lnot\text{correct\_large})$ -- how
often the small model is right where the large model is wrong -- not the accuracy
difference between the models. On PopQA that quantity was zero. This is a cheap pre-flight
computation and it invalidates entire experiments.

\paragraph{Cost the sampling overhead honestly.} Normalizing the $k$-sample pass to one
unit hides that entropy routing pays $k$ generations on every query before deciding
anything. Under per-generation accounting the method saves money only when
$\gamma > k/(1-r)$, which at $k=10$ demands a large model more than $11\times$ the
per-generation cost of the small one even at a $10\%$ escalation rate. Our GSM8K
configuration wins on accuracy and loses on cost -- both worth reporting, and the second
invisible under the normalized model.

\paragraph{On generalizing across datasets.}
\label{sec:discussion-generalising}
Three of our findings reversed between
benchmarks, which is the most uncomfortable result in this paper. Token confidence was below
chance on synthetic arithmetic and above it on GSM8K; fusing it degraded entropy in the first
case and improved Target 1 in the second; and retrieval-based prediction of large-model
success went from $0.908$ to chance. In each case the single-benchmark result would have
been reported as a property of the signal. We now think it should not be: a signal's
behavior here is a joint property of the signal and the benchmark's structure, and claims
of the form ``signal $X$ predicts $Y$'' should be accompanied by the controls that establish
which of the two is doing the work.

The retrieval case is the one where we can say what the benchmark contributed. Its
difficulty was lexically recoverable, embedding similarity therefore encoded difficulty, and
neighbor voting reduced to difficulty-conditioned outcome frequency -- an estimator our own
difficulty-calibrated baseline computes without any embedding. That the probe forecast the
collapse before we ran GSM8K is, to us, the clearest argument for reporting these controls
by default.

\paragraph{Limitations.} Our positive result rests on one dataset and one model pair. It
does replicate across two disjoint samples of that dataset ($n=250$ and $n=150$, agreeing to
$0.008$ AUROC on Target 2), but two samples from GSM8K are not two datasets, and nothing
here establishes that the result transfers to another task family or another model pair. Sampling temperature was fixed at $0.7$
throughout and never ablated, despite directly setting the diversity that semantic entropy
measures. We ablated the sample budget $k$ (Section~\ref{sec:kablation}) but not the
NLI entailment threshold $\tau$, which was fixed at $0.5$ throughout.  The knowledge-base result of
Section~\ref{sec:kb} additionally depends on the profiling and evaluation sets being
disjoint, which we verify by question-string comparison but which is weaker than sampling
them independently. The knowledge-base AUROC figures in Table~\ref{tab:kb-targets} and
Table~\ref{tab:tradeoff}, including the $0.908 \to 0.518$ collapse central to our retrieval
discussion, do not carry bootstrap confidence intervals: the per-item scores from those
single-run builds were not preserved, so an interval cannot be computed after the fact. This
is to be determined rather than resolved -- rerunning the knowledge-base profiling with
per-item scores preserved, specifically so a bootstrap interval can be computed, is the
concrete next step and is listed as such in Section~\ref{sec:future}. Finally, the failure
modes cataloged here are the ones we happened to detect; the reason we present them as a
methodological contribution is precisely that each was, for some period, invisible in
results that looked reasonable.

\section*{Acknowledgments}

This report synthesizes an iterative empirical investigation conducted across two
model pairs (Qwen2.5, Gemma-3), three reported benchmarks (synthetic arithmetic, GSM8K,
PopQA) plus TriviaQA and TruthfulQA loaders implemented but not run for this study, and two
independent codebases (a single-pass evaluation harness and a train/cal/test iterative
research loop).

\end{document}